\documentclass[conference]{IEEEtran}
\IEEEoverridecommandlockouts
\usepackage{cite}
\usepackage{amsmath,amssymb,amsfonts}
\usepackage{graphicx}
\usepackage{textcomp}
\usepackage[ruled,linesnumbered,vlined]{algorithm2e}

\usepackage{textcomp}
\usepackage{stfloats}
\usepackage[hyphens]{url}
\usepackage{verbatim}
\usepackage{graphicx}
\usepackage{cite}

\usepackage{amsmath}
\usepackage{amsfonts}

\usepackage{algpseudocode}
\usepackage{pifont}
\usepackage{color}
\usepackage{soul}
\usepackage{booktabs}
\usepackage{tabularx}
\usepackage{flushend}
\usepackage[normalem]{ulem}
\usepackage{pifont}
\usepackage{multirow}
\usepackage[table,xcdraw]{xcolor}
\usepackage{nicematrix}
\usepackage{array}
\usepackage{threeparttable}

\usepackage[colorlinks=true,
            urlcolor=black,
            linkcolor=black,
            anchorcolor=black,
            citecolor=black]{hyperref}

\usepackage{makecell}
\usepackage{multirow}
\usepackage{nicematrix}
\usepackage{float}
\usepackage[numbers,sort&compress]{natbib}
\usepackage[justification=justified]{caption}
\usepackage{subfigure}
\def\BibTeX{{\rm B\kern-.05em{\sc i\kern-.025em b}\kern-.08em
    T\kern-.1667em\lower.7ex\hbox{E}\kern-.125emX}}
\begin{document}

\title{PolyLUT-Add: FPGA-based \\LUT Inference with Wide Inputs
}

\author{\IEEEauthorblockN{Binglei Lou, Richard Rademacher, David Boland and Philip H.W. Leong}
\IEEEauthorblockA{School of Electrical and Computer Engineering\\
The University of Sydney, Australia 2006\\
Email: \{binglei.lou,richard.rademacher,david.boland,philip.leong\}@sydney.edu.au}
}

\maketitle

\begin{abstract}
FPGAs have distinct advantages as a technology for deploying deep neural networks (DNNs) at the edge. Lookup Table (LUT) based networks, where neurons are directly modeled using LUTs, help maximize this promise of offering ultra-low latency and high area efficiency on FPGAs. Unfortunately, LUT resource usage scales exponentially with the number of inputs to the LUT, restricting PolyLUT to small LUT sizes. This work introduces PolyLUT-Add, a technique that enhances neuron connectivity by combining $A$ PolyLUT sub-neurons via addition to improve accuracy. Moreover, we describe a novel architecture to improve its scalability. We evaluated our implementation over the MNIST, Jet Substructure classification, and Network Intrusion Detection benchmark and found that for similar accuracy, PolyLUT-Add achieves a LUT reduction of $2.0-13.9\times$ with a $1.2-1.6\times$ decrease in latency. 
\end{abstract}


\section{Introduction}

Deep neural networks (DNNs) have been shown to provide powerful feature extraction and regression capabilities and are widely employed across a spectrum of applications, including image classification for autonomous driving~\cite{lin2018architectural}, data analysis in particle physics~\cite{duarte2018fast}, and real-time anomaly detection~\cite{murovic2019massively,fsead}. Field-Programmable Gate Arrays (FPGAs) provide a unique implementation platform for deploying DNNs, with significant advantages over other technologies, particularly in real-time inference tasks.

Lookup Table (LUT) based neurons on FPGAs offer high area efficiency and ultra-low latency. Examples of accelerators published using this approach include LUTNet~\cite{LUTNet}, NullaNet~\cite{NullaNet}, LogicNets~\cite{LogicNets}, and PolyLUT~\cite{polylut}. Compared with Binary Neural Networks (BNNs~\cite{bnn}), which utilize 1-bit quantization to replace multipliers with simple XNOR gates, LUT-based neurons further optimize FPGA resource utilization using LUTs as direct inference operators. 

\begin{figure*}[]
    \centerline{\includegraphics[width=1.0\linewidth]{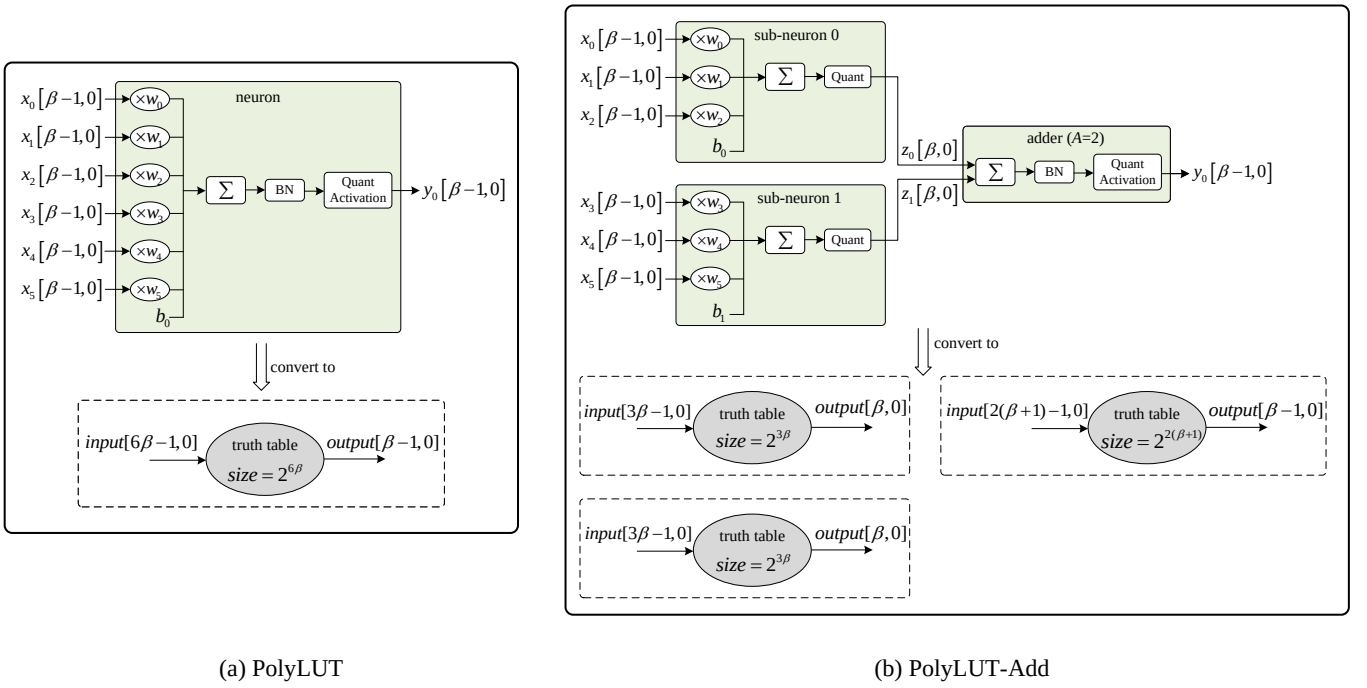}}
    \caption{Architecture of a (a) PolyLUT and (b) PolyLUT-Add neuron. Fan-in $F$ is 6 (each of $\beta$-bit words), and the sub-neuron number $A$ is set to 2. For simplicity, the polynomial order of each PolyLUT neuron~\cite{polylut} is set to 1 in this example. For each output bit, PolyLUT requires a number of lookup table entries of $2^{6 \beta}$, while PolyLUT-Add requires ($2^{3\beta}$ + $2^{3 \beta}$ + $2^{2 (\beta+1)}$).}
    \label{fg:polylut}
\end{figure*}

Building upon the PolyLUT framework, this work introduces an enhancement called PolyLUT-Add, where we combine $A$ copies of PolyLUT sub-neurons via an $A$-input adder to increase neuron fan-in. 
Figure~\ref{fg:polylut} highlights how our approach builds from PolyLUT for a simple example where $A=2$. The computation process of a PolyLUT neuron: weight multiplication, accumulation, batch normalization (BN), and quantized activation, is first shown in Figure~\ref{fg:polylut}(a). Our PolyLUT-Add approach, shown in Figure~\ref{fg:polylut}(b), restructures the neuron computation. The first stage is similar to PolyLUT without the batch normalization and is repeated for each sub-neuron. Instead, the batch normalization is performed after the results are accumulated, with the resulting activation quantized again if necessary.

While the same functionality of PolyLUT-Add could be achieved with PolyLUT with a single lookup table, PolyLUT-Add can make better use of the FPGA fabric. In this example, with the input word length, $\beta=2$, PolyLUT-Add uses three distinct lookup tables, each of size $2^{6}$ (the number of entries); the single lookup table equivalent would be of size $2^{12}$. In the general case, if we define the fan-in to be $F$ of ($\beta$-bit words), for each output bit, PolyLUT requires a lookup table of $\mathcal{O}(2^{\beta FA})$, while PolyLUT-Add reducing it to $\mathcal{O}(A \times 2^{\beta F} + 2^{A(\beta +1)})$.

The contributions of this work can be summarized as:

\begin{itemize}

\item At the algorithmic level, we introduce PolyLUT-Add, an extension of the PolyLUT framework~\cite{polylut}, which incorporates $A$ PolyLUT sub-neurons, combined via an $A$-input adder to enable improved accuracy.
\item At the computer architecture level, we propose an efficient FPGA implementation of PolyLUT-Add.
\item To the best of our knowledge, for similar accuracy, PolyLUT-Add produces the best-reported FPGA latency and area results on the three datasets tested. The design has been made open source on GitHub to facilitate reproducible research~\footnote{PolyLUT-Add: \url{https://github.com/bingleilou/PolyLUT-Add}}.  
\end{itemize}

We evaluated PolyLUT-Add across three datasets with four DNN models and demonstrated significant accuracy improvements. Specifically, for the same polynomial degree $D$ and fan-in $F$ setups, $A=2$ achieves an accuracy improvement of up to 2.7\%, albeit with a 2-3 fold increase in area. Latency and clock frequency are unchanged in most cases. 
However, we also see that when $A=2$, we can choose a lower $D$ and $F$ and obtain accuracy levels comparable to those achieved by the original PolyLUT.
This reduces LUT consumption by factors of 4.8, 13.9, and 2.0 for the MNIST, Jet Substructure classification, and Network Intrusion Detection benchmarks, respectively, with a 1.2 to 1.6 times decrease in latency.

The remainder of this paper is organized as follows. In Section~\ref{se:background}, we review previous work on LUT-based neurons. In Section~\ref{se:design}, the design of PolyLUT-Add is described. Results are presented in Section~\ref{se:results} and conclusions drawn in Section~\ref{se:conclusion}.

\section{Background}
\label{se:background}

Wang {\em et al.} introduced LUTNet, the first LUT-optimized FPGA inference scheme~\cite{LUTNet}. Its approach was to prune a residual BNN: ReBNet~\cite{ReBNet} by mapping some of the XNOR-population count (popcount) operations directly to $k-$input LUTs to take advantage of FPGA architectures.
NullaNet~\cite{NullaNet} and LogicNets~\cite{LogicNets} and adopted a different approach by quantizing the inputs and outputs of each neuron and encapsulating the neuron's transfer function ({\em i.e.}, densely connected linear and activation functions) in a lookup table. This method enumerated all possible combinations of a neuron's inputs and determined the corresponding outputs based on the neuron's weights and biases. By replacing popcount operations with Boolean expressions, significant computational savings were made. Building upon the foundations of LogicNets, PolyLUT~\cite{polylut}, proposed by Andronic {\em et al.}, further enhanced accuracy and reduced the number of required layers by introducing piecewise polynomial functions. 

\begin{figure}
    \centerline{\includegraphics[width=1.0\linewidth]{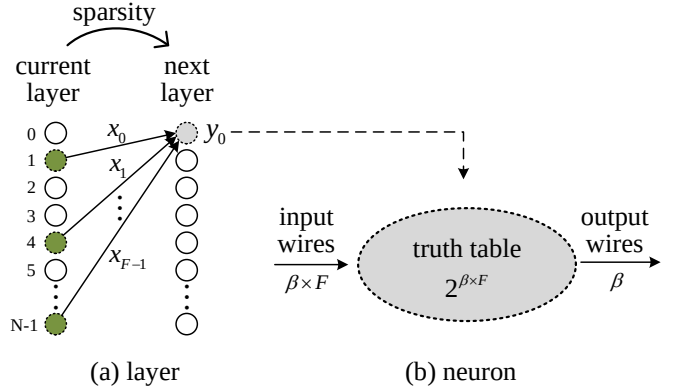}}
    \caption{Illustration of the LUT-based DNN inference scheme used in LogicNets~\cite{LogicNets} and PolyLUT~\cite{polylut}.} 
    \label{fg:lut}
\end{figure}

Figure~\ref{fg:lut} illustrates the main idea behind LogicNets and PolyLUT approaches. Only sparse connections of maximum $F$ inputs from the previous layer are supported, and these can be directly mapped to the output via LUTs, eliminating the $F$-bit popcount operations required to form the sum for a dot product operation. As an example in Figure~\ref{fg:lut}(a), the current layer has $N$ neurons, of which only $F$ random nodes are used as inputs to each neuron in the next layer. In Figure~\ref{fg:lut}(b), the transfer function mapping an input vector $\left[ {{x_0},{x_1}, \ldots ,{x_{F - 1}}} \right]$ to the output node ${y_0}$ can be implemented using $\beta F$ inputs, and hence requires lookup table entries of $2^{\beta F}$.
The constraint $F \ll N$ is manually applied to limit the number of the lookup table entries.

\begin{small}
\begin{equation}
    \label{eq:poly}
    {y_0} = \sigma \left( {\sum\limits_{i = 0}^{M - 1} {{w_i}{m_i}\left( x \right) + b} } \right),{\text{where }}M = \left( \begin{gathered}
  F + D \\ 
  D \\ 
\end{gathered}  \right)
\end{equation}
\end{small}

The computation inside the neuron ${y_0}$ can be described as Eq.~\eqref{eq:poly}, where $\sigma$ is the quantized activation function, $w$ and $b$ denote weight and bias respectively, $D$ is a polynomial degree. For the LogicNets method, $D$ is 1; PolyLUT generalizes these methods by allowing larger polynomial degrees constructed from multiplicative combinations of the inputs up to degree $D$. For instance, if the input vector is two-dimensional and $D=2$, the model construction proceeds as follows:
$\left[ {{x_0},{x_1}} \right] \to \left[ {1,{x_0},{x_1},x_0^2,{x_0}{x_1},x_1^2} \right]$. The value of $M$ equals the number of monomials $m(x)$ of at most degree $D$ in $F$ variables.

While PolyLUT enriched the representational capability of previous solutions, this requires a lookup table of $\mathcal{O}(2^{\beta F})$.
For instance, in the context of the MNIST handwritten digit recognition task~\cite{mnist}, which involves classifying 28$\times$28 pixel images into 10 categories, PolyLUT's architecture employs layers with (784, 256, 100, 100, 100, 10) neurons, using parameters $\beta =2$ and $F=6$. This means that only 6 neurons are randomly selected from each layer to form extremely sparse connectivity to neurons in the next layer. To address scalability by avoiding very large table sizes, the number of each neuron's lookup table entries was capped at $2^{12}$.
The exponential LUT requirement of this approach precludes the selection of larger $\beta$ and fan-in values, which can in turn limit accuracy.

\section{Design}
\label{se:design}

\subsection{DNN architecture}
\begin{figure}
    \centerline{\includegraphics[width=1.0\linewidth]{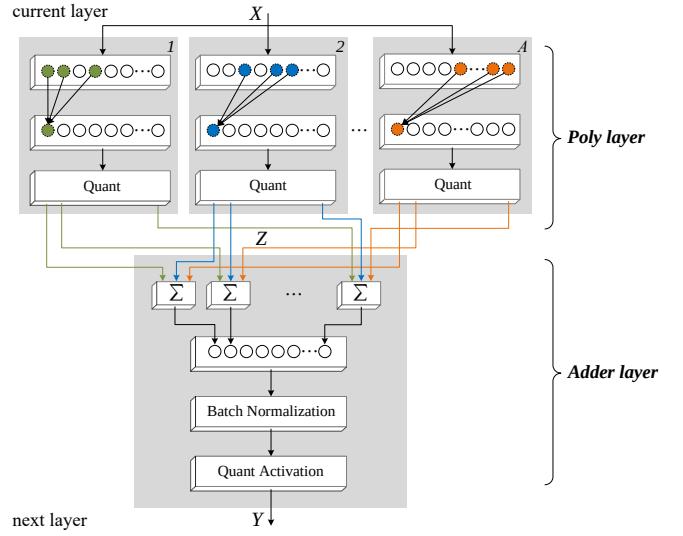}}
    \caption{A single-layer block diagram of PolyLUT-Add.}
    \label{fg:layer_structure}
\end{figure}

Figure~\ref{fg:layer_structure} outlines our proposed DNN architecture. Compared with Figure~\ref{fg:lut}, the fan-in $F$ to sub-neurons remains the same, but the total fan-in to the neuron is increased by a factor of $A$ at the output. This is achieved by summing $A$ independent and parallel randomly connected Poly-layers. 
To elucidate the enhancement mechanism, we introduce the formulation detailed in Eq.~\eqref{eq:equal}.

\begin{small}
\begin{equation}
    \label{eq:equal}
    \sum\limits_{i = 0}^{AF - 1} {{w_i}{x_i} + b}  = \sum\limits_{a = 0}^{A - 1} {\left( {\sum\limits_{i = 0}^{F - 1} {{w_{(aF + i)}}{x_{(aF + i)}}}  + {b_a}} \right)} 
\end{equation}
\end{small}

During computation, the activation function, such as Rectified linear unit (ReLU) output bits, can be one bit less than the input bits because its output is non-negative. To avoid overflow in the Adder-layer, we increase its internal word length by one bit (to $\beta+1$), as seen in Figure~\ref{fg:polylut}(b).

\subsection{System Toolflow}

\begin{figure}
\centerline{\includegraphics[width=0.86\linewidth]{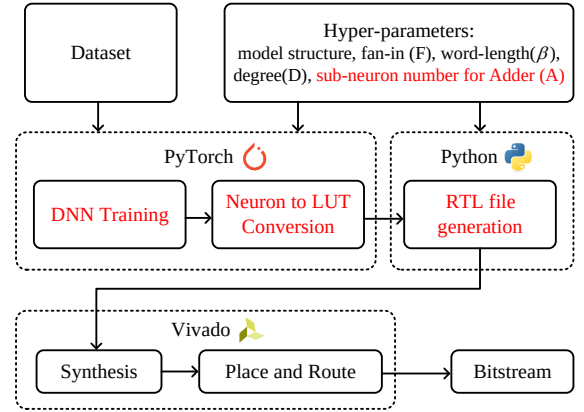}}
\caption{Tool flow for PolyLUT-Add. The original open-source PolyLUT toolflow~\cite{polylut} components are shown in black, with modified elements in red.}
\label{fg:toolflow}
\end{figure}

Figure~\ref{fg:toolflow} shows the tool flow. Like PolyLUT, training is done offline using PyTorch~\cite{PyTorch}, and then the resulting weights are used to create the LUTs that implement each neuron. These tables are then utilized to generate Register Transfer Level (RTL) files in Verilog, encapsulating the Boolean expressions derived from the neurons. The final stage involves synthesizing the LUT-based DNN design onto hardware, using the AMD/Xilinx Vivado tool~\cite{Vivado}.

The integration of Brevitas~\cite{brevitas} with PyTorch facilitates quantization-aware training of DNNs. We modified the network implementation to accept $A$, the model's fan-in factor as a parameter. The model's weights are transformed into lookup tables following the training phase. This transformation begins by employing the quantized states inherent in the trained model to ascertain each neuron's input data range. For Poly-layers, we generate all possible input combinations based on $\beta$ and $F$; In contrast, for the Adder-layer, all combinations are generated based on $\beta$ and $A$. These input combinations are subsequently fed into their respective layers---Poly-layer and Adder-layer---to generate the corresponding outputs. Finally, these input and output pairs form the individual values for the lookup table.

\subsection{Pipelining}
\label{se:Latency}

\begin{figure}
    \centerline{\includegraphics[width=0.9\linewidth]{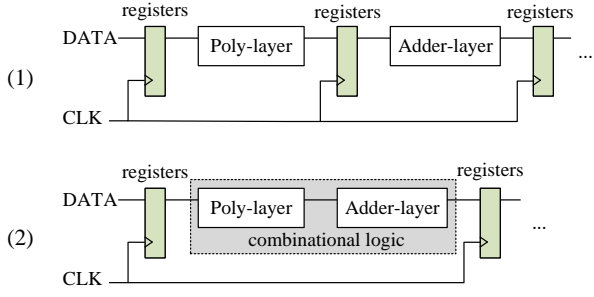}}
    \caption{Two synthesis strategies}
    \label{fg:clk}
\end{figure}

In our FPGA design, we treat each layer as an independent module and synthesize them separately, with the critical path in the layer with the largest delay determining the system's maximum clock frequency. Two implementation strategies were considered, as illustrated in Figure~\ref{fg:clk}.

\begin{enumerate}
\item 
\textbf{Separate pipeline registers for each layer}. This strategy is best when the lookup table size (a proxy for critical path delay) of the Adder-layer ($\mathcal{O}(2^{A(\beta +1)})$) and Poly-layer ($\mathcal{O}(A 2^{\beta F})$) are similar. Although the latency in clock cycles is doubled, this architecture maximizes clock frequency.

\item 
\textbf{Single register for combined Poly-layer and Adder-layer}. When the Adder-layer is much smaller than the Poly-layer, its processing time should not adversely impact the Poly-layer's performance, enabling a more efficient overall system design where latency is tightly controlled.
\end{enumerate}

\section{Results}
\label{se:results}

\subsection{Datasets}
We evaluated the proposed PolyLUT-Add design on three commonly used datasets for ultra-low latency inference:

\begin{table}[!b]
\centering
  \caption{Model setups used to evaluate different datasets.}
  \def\arraystretch{0.8}
  \scalebox{0.82}{
  \begin{tabular}{ccccccc}
  \toprule
  Dataset & Model name & Neurons per layer & $\beta$ & $F$ & $D$ & $A$   \\
  \midrule
  MNIST            & HDR            & 256, 100, 100, 100, 100, 10 & 2 & 6 & 1, 2 & 2, 3    \\
  \midrule
  Jet Substructure & JSC-XL{\ding{172}}     & 128, 64, 64, 64, 5          & 5 & 3 & 1, 2 & 2      \\
  \midrule
  Jet Substructure & JSC-M Lite     & 64, 32, 5                   & 3 & 4 & 1, 2 & 2, 3   \\
  \midrule
  UNSW-NB15   & NID Lite{\ding{173}}   & 686, 147, 98, 49, 1         & 3 & 5 & 1    & 2      \\
  \bottomrule
\end{tabular}}
\label{tb:setup}

\begin{tablenotes}
    \footnotesize
    \item[1] Remarks: {\ding{172}} $\beta_{i}=7$, $F_{i}$ = 2;   \quad\quad\quad {\ding{173}} $\beta_{i}=1$, $F_{i}$ = 7;
\end{tablenotes}

\end{table}

\begin{enumerate}
\item \textit{Handwritten Digit Recognition}: In timing-critical sectors such as autonomous vehicles, medical imaging, and real-time object tracking, the demand for low-latency image classification is paramount. These applications underscore the necessity for swift and accurate decision-making, where even minimal delays can have significant repercussions. Unfortunately, there is no public dataset specialized for low-latency image classification tasks. The MNIST~\cite{mnist} is therefore utilized to benchmark our work on its image classification performance, which is a dataset for handwritten digit recognition tasks with $28 \times 28$ pixels as input image and 10 classes as outputs.

\item \textit{Jet Substructure Classification}: Real-time decision-making is often important for physics experiments such as the CERN Large Hadron Collider (LHC). Jet Substructure Classification (JSC) is one of its applications that requires high-throughput data processing. Prior works~\cite{ngadiuba2020compressing,duarte2018fast,coelho2021automatic,fahim2021hls4ml} employed neural networks on FPGA for this task to provide real-time inference capabilities. We also use the JSC dataset formulated from Ref.~\cite{duarte2018fast} to evaluate our work, with the dataset having 16 substructure properties as input and 5 types of jets as outputs.

\item \textit{Network Intrusion Detection}: In the field of cybersecurity, the swift detection and mitigation of network threats are important for the preservation of digital infrastructure integrity ({\em e.g.}, fiber-optic throughput can reach 940 Mbps). Prior works have used FPGAs to accelerate DNNs, enabling real-time Network Intrusion Detection Systems (NIDS) with high accuracy and enabling privacy on edge devices~\cite{polylut,LogicNets,murovivc2021genetically}. The UNSW-NB15 dataset~\cite{unsw} was used as the benchmark for our evaluation process. It has 49 input features and binary classification (bad or normal).

\end{enumerate}

Results for the JSC and NID were reported in the LogicNets paper~\cite{LogicNets}, and all datasets were used for PolyLUT~\cite{polylut}.

\begin{figure}[!b]
    \centering
    \subfigure[HDR]{
    \hspace{-0.05\linewidth}
        \centering
        \label{fg:mnist}
        \includegraphics[width=0.89\linewidth]{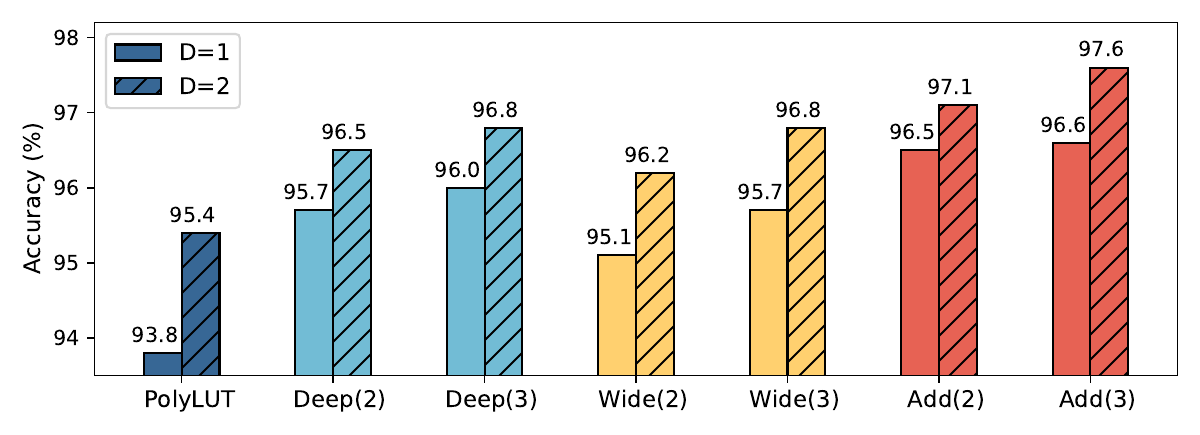}
    }\vspace{-2.3mm}
    \subfigure[JSC-XL]{
    \hspace{-0.05\linewidth}
        \label{fg:jsc-xl}
        \includegraphics[width=0.5\linewidth]{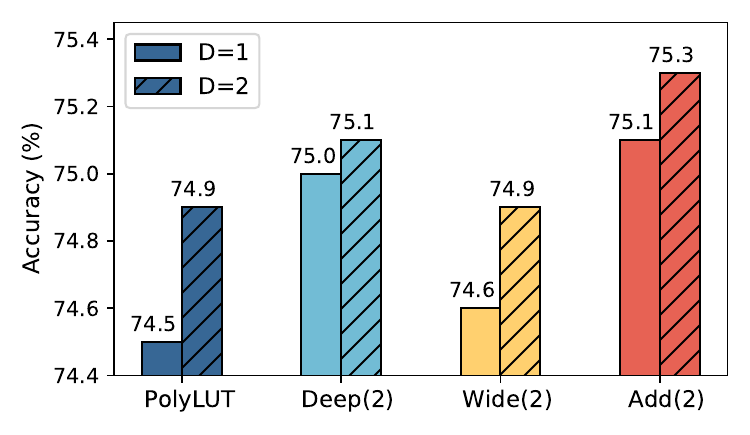}
    }\vspace{-2.3mm}
    
    \subfigure[JSC-M Lite]{
    \hspace{-0.05\linewidth}
        \centering
        \label{fg:jsc-mlite}
        \includegraphics[width=0.89\linewidth]{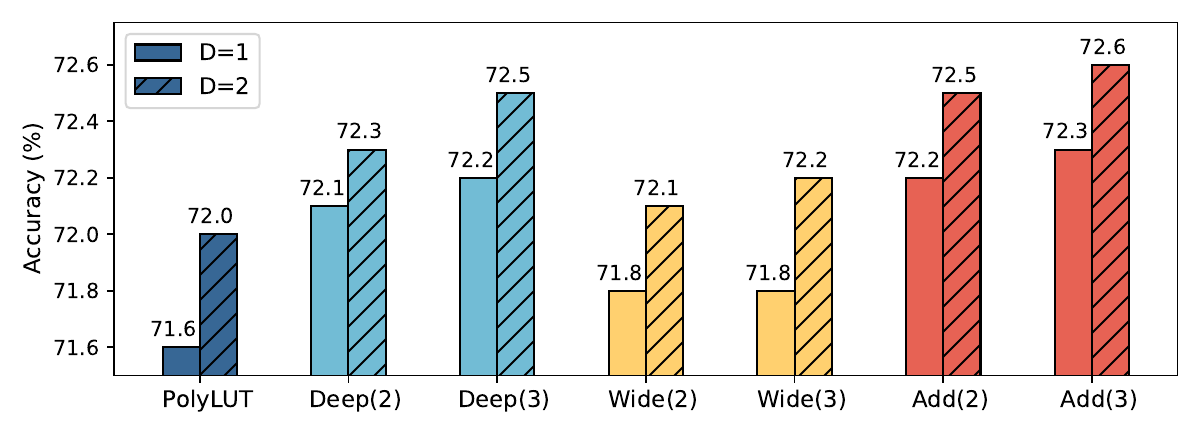}
    }\vspace{-2.3mm}
    \caption{Accuracy results on different models. We use Deep($\mathbb{D}$), Wide($\mathbb{W}$) and Add($A$) to denote ``PolyLUT-Deeper'', ``PolyLUT-Wider'' and ``PolyLUT-Add'' respectively.}
    \label{fg:acc_results}
\end{figure}
\subsection{Experimental Setup}

Table~\ref{tb:setup} lists our neural network configurations for three datasets. As a foundation for our experiments and to ensure consistency in evaluation, our setup closely follows the PolyLUT study~\cite{polylut}. Our newly introduced $A$ is set to $A \in \{2,3\}$ for models (HDR and JSC-M Lite) with small truth table ($\mathcal{O}(2^{\beta F})$), and $A=2$ is used for JSC-XL. Larger $A$ ($A\geq4$) can be supported using an adder tree, which is left as future work.
 
The polynomial degree $D=1$ and $D=2$ correspond to linear and quadratic representations, respectively. 
We utilize $D \in \{1,2\}$ to evaluate the performance of PolyLUT-Add.
As will be seen later, to facilitate a comparison with existing literature and aim for enhanced accuracy, we also explore $D=3$ in Section~\ref{se:area}. This contrasts with PolyLUT, for which higher degrees ($D \in \{4,6\}$) achieve the best accuracy.
We comment the case $A=1$ is identical to PolyLUT, and $A=1, D=1$ corresponds to LogicNets. We also note that the training convergence of the UNSW dataset is sensitive to the initial random seed, and hence, multiple trials were necessary before a result with good accuracy was achieved. As an exception, we therefore apply $A=2$ and $D=1$ to evaluate the NID Lite model.

Using AdamW as the optimizer~\cite{PyTorch}, we trained the smaller models/datasets (JSC-M Lite, NID Lite) for 1000 epochs and used 500 epochs for (JSC-XL and HDR). The mini-batch size is set to 1024 and 128 for (JSC-XL, JSC-M Lite, NID-Lite) and MNIST, respectively. 
We inherit these configurations from PolyLUT~\cite{polylut} to ensure consistency in evaluation.

\begin{table*}[!b]
  \centering
  \caption{Comparison of accuracy and hardware results between PolyLUT and PolyLUT-Add ($\mathbb{D}=1$, $\mathbb{W}=1$)}
  \label{tb:resuts_3datasets}
  \bgroup
  \def\arraystretch{1.2}
  \setlength\tabcolsep{3mm}
  \scalebox{0.85}{
  \begin{tabular}{c|c|l|c|c|l|cccc|c}
  \cline{1-11}
  \multirow{2}{*}{\textbf{Models}}  & \multirow{1}{*}{\textbf{Degree}} & \multicolumn{1}{c|}{\multirow{2}{*}{\textbf{Model}}} & \textbf{Fan-in} & \multirow{2}{*}{\textbf{Acc(\%)$\uparrow$}} & \multicolumn{1}{c|}{ \textbf{lookup table}} & \textbf{LUT}          & \textbf{FF} & \textbf{$F\_max$} & \textbf{Latency}  & \textbf{RTL Gen.}                 \\ 
   & $D$ & & \textbf{$ (F \times A)$} & & \multicolumn{1}{c|}{ \textbf{entries$\downarrow$} }    &  \textbf{(\% of 1182240)}          & \textbf{(\% of 2364480)} & \textbf{(MHz)} & \textbf{(cycles)}   & \textbf{(hours)} \\ \hline\hline
   \multirow{8}{*}{HDR}  & \multirow{4}{*}{1} & \multirow{2}{*}{PolyLUT}        & 6          & 93.8       & $2^{12}$                     & 3.43& 0.12 & 378 & 6 & 1.40  \\
                           &                    &                                  & 10         & 96.1      & $2^{12} \times 256$          & \multicolumn{4}{c|}{ \(-\) } & \(-\)  \\ \cline{3-3}
                           &                    & \multirow{2}{*}{PolyLUT-Add}     & 6$\times$2 & 96.5       & $2^{12} \times 2$ + $2^{6}$  & 12.69 & 0.12 & 378 & 6 & 3.00  \\
                           &                    &                                  & 6$\times$3 & \bf{96.6}  & $2^{12} \times 3$ + $2^{9}$  & 20.67 & 0.12 & 378 & 6 & 4.40  \\ \cline{2-11}
                           & \multirow{4}{*}{2} & \multirow{2}{*}{PolyLUT}         & 6          & 95.4       & $2^{12}$                     & 6.62 & 0.12 & 378 & 6 & 1.40  \\
                           &                    &                                  & 10         & 97.3       & $2^{12} \times 256$           & \multicolumn{4}{c|}{\(-\)} & \(-\) \\ \cline{3-3}
                           &                    & \multirow{2}{*}{PolyLUT-Add}     & 6$\times$2 & 97.1       & $2^{12} \times 2$ + $2^{6}$  & 19.78 & 0.07 & 378 & 6 & 3.00  \\
                           &                    &                                  & 6$\times$3 & \bf{97.6}  & $2^{12} \times 3$ + $2^{9}$  & 31.36 & 0.07 & 378 & 6 & 4.50  \\ \hline\hline
  
   \multirow{6}{*}{JSC-XL}  & \multirow{3}{*}{1} & \multirow{2}{*}{PolyLUT}       & 3          & 74.5       & $2^{15}$                  & 19.55& 0.07& 235 & 5 & 2.10  \\
                           &                    &                                  & 5          & 74.9       & $2^{15} \times 1024$        & \multicolumn{4}{c|}{ \(-\) } & \(-\)  \\ \cline{3-3}
                           &                    & \multirow{1}{*}{PolyLUT-Add}     & 3$\times$2 & \bf{75.1}  & $2^{15} \times 2$ + $2^{12}$ & 50.10 & 0.07 & 235 & 5 & 5.17  \\ \cline{2-11}
                           & \multirow{3}{*}{2} & \multirow{2}{*}{PolyLUT}         & 3          & 74.9       & $2^{15}$                     & 37.40 & 0.07 & 235 & 5 & 2.30  \\
                           &                    &                                  & 5          & 75.2       & $2^{15} \times 1024$         & \multicolumn{4}{c|}{ \(-\) } & \(-\)  \\ \cline{3-3}
                           &                    & \multirow{1}{*}{PolyLUT-Add}     & 3$\times$2 & \bf{75.3}  & $2^{15} \times 2$ + $2^{12}$ & 89.60 & 0.07 & 235 & 5 & 5.24  \\ \hline\hline

   \multirow{8}{*}{JSC-M Lite}  & \multirow{4}{*}{1} & \multirow{2}{*}{PolyLUT}   & 4          & 71.6       & $2^{12}$                     & 0.97 & 0.01 & 646 & 3 & 0.16  \\
                           &                    &                                  & 7          & 72.1       & $2^{12} \times 512$            & \multicolumn{4}{c|}{ \(-\) } & \(-\)  \\ \cline{3-3}
                           &                    & \multirow{2}{*}{PolyLUT-Add}     & 4$\times$2 & 72.2       & $2^{12} \times 2$ + $2^{8}$  & 2.62 & 0.01 & 488 & 3 & 0.35  \\
                           &                    &                                  & 4$\times$3 & \bf{72.3}  & $2^{12} \times 3$ + $2^{12}$ & 4.33& 0.01& 363 & 3 & 0.63  \\ \cline{2-11}
                           & \multirow{4}{*}{2} & \multirow{2}{*}{PolyLUT}         & 4          & 72.0       & $2^{12}$                     & 1.51& 0.01& 568 & 3 & 0.16  \\
                           &                    &                                  & 6          & 72.5       & $2^{12} \times 512$          & \multicolumn{4}{c|}{ \(-\) } & \(-\)  \\ \cline{3-3}
                           &                    & \multirow{2}{*}{PolyLUT-Add}     & 4$\times$2 & 72.5       & $2^{12} \times 2$ + $2^{8}$  & 4.29& 0.01& 440 & 3 & 0.34  \\
                           &                    &                                  & 4$\times$3 & \bf{72.6}  & $2^{12} \times 3$ + $2^{12}$ & 6.57& 0.01& 373 & 3 & 0.64  \\ \hline\hline
    \multirow{3}{*}{NID Lite}  & \multirow{3}{*}{1} & \multirow{2}{*}{PolyLUT}      & 5          & 89.3       & $2^{15}$                     & 6.86     & 0.15   & 529 & 5  & 4.09  \\
                           &                    &                                  & 8          & 91.0       & $2^{15} \times 512$          & \multicolumn{4}{c|}{ \(-\) } & \(-\)  \\ \cline{3-3}
                           &                    & PolyLUT-Add                      & 5$\times$2 & \bf{91.6}  & $2^{15} \times 2$ + $2^{8}$ & 21.41    & 0.15    & 529 & 5  & 8.76  \\ \cline{2-11}\hline  
  \end{tabular}}
  \egroup

  \begin{tablenotes}
    \footnotesize
    \item[1] \(-\): Data for very high fan-in settings is omitted due to exceeding FPGA memory capacity limits.
\end{tablenotes}
\end{table*}

We used the AMD/Xilinx \texttt{xcvu9p-flgb2104-2-i} FPGA part for evaluation to facilitate comparisons with PolyLUT~\cite{polylut} and LogicNets~\cite{LogicNets}. The designs are compiled using Vivado 2020.1 with \texttt{Flow\_PerfOptimized\_high} settings and are configured to perform synthesis in the \texttt{Out-of-Context} (\texttt{OOC}) mode.
The RTL Generation time was measured on a desktop with Intel(R) Core(TM) i7-10700F @2.9GHz and 64GB memory.

\subsection{PolyLUT-Add vs. Deeper and Wider PolyLUT (small $D$)}
\label{se:Accuracy}

We first present results comparing PolyLUT-Add with PolyLUT in configurations with the same polynomial degree. Three configurations were tested:
\begin{enumerate}
\item \textbf{Original PolyLUT}: This serves as the baseline network.

\item \textbf{PolyLUT-Deeper}: This explores the impact of increasing network depth. We denote the depth factor as $\mathbb{D}$. Then $\mathbb{D}\times$ the number of layers is applied to models in Table~\ref{tb:setup}. For example, for JSC-M Lite, if $\mathbb{D}=2$, the hidden layer is doubled, meaning the neurons per layer becomes (64,64,32,32,5). 

\item \textbf{PolyLUT-Wider}: This examines the impact of a wider network model. We denote the width factor as $\mathbb{W}$. Then $\mathbb{W}\times$ the number of neurons per layer are applied to models in Table~\ref{tb:setup}. Once again, for JSC-M Lite, if $\mathbb{W}=2$, the neurons per layer becomes (128,64,5).
\end{enumerate}

Figure~\ref{fg:acc_results} shows the accuracy of the configurations with parameter settings detailed in Table~\ref{tb:setup}. PolyLUT-Add achieves the highest accuracy against all baselines on all datasets for both the linear ($D=1$) and non-linear ($D=2$) cases.

\subsection{Optimizing for Accuracy}
\label{se:area}

\begin{table*}[]
  \centering
  \caption{Comparison results with prior works. PolyLUT-Add uses smaller $F$ and $D$ (see Table~\ref{tb:setup_small}), whereas PolyLUT uses larger $D$, $F$ for accuracy. The frequency and the area are collected from the Vivado post Place \& Route reports. }
  \label{tb:comparison}
  \bgroup
  \def\arraystretch{1.20}
  \setlength\tabcolsep{3mm}
  \scalebox{0.91}{
  \begin{tabular}{c|lccccccc}
  \cline{1-9}
  Dataset                        & Model                      & Accuracy$\uparrow$  & LUT    & FF     & DSP & BRAM & $F\_max$(MHz)$\uparrow$ & Latency(ns)$\downarrow$ \\ \hline
  \multirow{4}{*}{MNIST}         & \bf{PolyLUT-Add (HDR-Add2, $D$=3)}              & \bf{96\%}   & \bf{14810}   &  \bf{2609}  & \bf{0}   & \bf{0}    & \bf{625}           &   \bf{10}     \\
                                 & PolyLUT (HDR, $D$=4)~\cite{polylut}        & \bf{96\%}   & 70673        & 4681        & \bf{0}   & \bf{0}    & 378           & 16       \\
                                 & FINN~\cite{finn}                           & \bf{96\%}   & 91131        &  -          & \bf{0}   & 5    & 200           & 310       \\
                                 & \texttt{hls4ml}~\cite{hls4ml}              & 95\%        & 260092       & 165513      & \bf{0}   & \bf{0}    & 200           & 190       \\\hline\hline
  \multirow{4}{*}{Jet Substructure}   & \bf{PolyLUT-Add (JSC-XL-Add2, $D$=3)}           & 75\%        &  \bf{36484}   &  \bf{1209}   & \bf{0}   & \bf{0}    & \bf{315}           & \bf{16}       \\
                                 & PolyLUT (JSC-XL, $D$=4)~\cite{polylut}     & 75\%        & 236541       & 2775   & \bf{0}   & \bf{0}    & 235           & 21       \\
                                 & Duarte {\em et al.}~\cite{duarte2018fast}  & 75\%        &  \multicolumn{2}{c}{88797$^*$} & 954   & \bf{0}    & 200           & 75       \\
                                 & Fahim {\em et al.}~\cite{fahim2021hls4ml}  & \bf{76\%}   & 63251        & 4394      & 38  & \bf{0}    & 200           & 45       \\\hline\hline
  \multirow{3}{*}{Jet Substructure}   & \bf{PolyLUT-Add (JSC-M Lite-Add2, $D$=3)}       & \bf{72\%}   &  \bf{895}   & \bf{189}  & \bf{0} & \bf{0}    & \bf{750}           & \bf{4}       \\
                                 & PolyLUT (JSC-M Lite, $D$=6)~\cite{polylut} & \bf{72\%}   & 12436        & 773       & \bf{0}   & \bf{0}    & 646           & 5       \\
                                 & LogicNets~\cite{LogicNets}                 & \bf{72\%}   & 37931        &  810      & \bf{0}   & \bf{0}    & 427           & 13       \\ \hline\hline
    \multirow{4}{*}{UNSW-NB15}     & \bf{PolyLUT-Add (NID-Add2, $D$=1)}                & \bf{92\%}    &  \bf{1649}   &  830     & \bf{0}   & \bf{0}    & \bf{620}           & \bf{8}       \\
                                 & PolyLUT (NID-Lite $D$=4)~\cite{polylut}     & \bf{92\%}    & 3336   & 686    & \bf{0}   & \bf{0}    & 529           & 9       \\
                                 & LogicNets~\cite{LogicNets} & 91\%      & 15949  & 1274   & \bf{0}   & 5    & 471           & 13       \\
                                 & Murovic {\em et al.}~\cite{murovivc2021genetically}      & \bf{92\%}      & 17990  & \bf{0}      & \bf{0}   & \bf{0}    & 55            & 18       \\\hline

  \end{tabular}}
  \egroup

  \begin{tablenotes}
    \footnotesize
    \item[1] \(*\): Paper reports ``LUT+FF''
\end{tablenotes}
\end{table*}

\begin{table}[]
\centering
  \caption{Model setups for smaller $F$ of PolyLUT-Add. }
  \def\arraystretch{0.8}
  \scalebox{0.8}{
  \begin{tabular}{ccccccc}
  \toprule
  Dataset          & Model name          & Neurons per layer           & $\beta$ & $F$ & $D$ & $A$  \\
  \midrule
  MNIST            & HDR-Add2           & 256, 100, 100, 100, 100, 10 & 2       & 4   & 3   & 2    \\
  \midrule
  Jet Substructure & JSC-XL-Add2{\ding{172}}    & 128, 64, 64, 64, 5          & 5       & 2   & 3   & 2  \\
  \midrule
  Jet Substructure & JSC-M Lite-Add2    & 64, 32, 5                   & 3       & 2   & 3   & 2   \\
  \midrule
  UNSW-NB15        &  NID-Add2{\ding{173}}      & 100, 100, 50, 50, 1         & 2       & 3   & 1   & 2   \\
  
  \bottomrule
\end{tabular}}
\label{tb:setup_small}
\begin{tablenotes}
    \footnotesize
    \item[1]  Remarks: {\ding{172}} $\beta_{i}=7$, $F_{i}$ = 1;  \quad\quad {\ding{173}} $\beta_{i}=1$, $F_{i}$ = 6, $\beta_{o}=2$, $F_{o}$ = 7
\end{tablenotes}
\end{table}

In terms of accuracy and hardware, Table~\ref{tb:resuts_3datasets} shows that for $A=2$, PolyLUT-Add achieved accuracy improvements of 2.7\%, 0.6\% and 2.3\% over PolyLUT on the MNIST, Jet Substructure classification and Network Intrusion Detection benchmarks respectively. However, this required a 2-3$\times$ increase in LUT size. 

We also evaluate the performance of simply increasing PolyLUT's fan-in, $F$. This has a number of lookup table entries of 256-1024$\times$ for similar accuracy, showing that PolyLUT-Add can improve model accuracy without an excessive impact on LUT size. Furthermore, it's noteworthy that the RTL Generation time cost also correlates with the number of lookup table entries; it follows that a direct increase in fan-in would incur exponentially higher RTL Generation time costs.

In terms of latency, we apply single registers for combined Poly-layer and Adder-layer (pipeline strategy-(2) in Figure~\ref{fg:clk}) to models in Table~\ref{tb:resuts_3datasets}. For HDR, JSC-XL, and NID Lite, PolyLUT-Add achieves the same latency (with maximum frequency ($F\_max$), which was constrained at 378 MHz, 235 MHz, and 529 MHz respectively in Ref.~\cite{polylut}). However, on the JSC-M Lite model, $F\_max$ is decreased. Therefore, we use the JSC-M Lite model as a case study to analyze its maximum frequency and clock cycles for pipeline strategies (1) and -(2). The results are shown in Table~\ref{tb:pipeline}. As expected, the separate pipeline registers for each layer (strategy-(1)) do not affect overall system performance, whereas strategy-(2) results in the lowest overall latency with lower $F\_max$. We suggest that the best choice will be dependent on specific system requirements.

\begin{table}[]
  \centering
  \caption{Comparison of two pipeline strategies on PolyLUT-Add with JSC-M Lite as the case study}
  \label{tb:pipeline}
  \bgroup
  \def\arraystretch{1.02}
  \setlength\tabcolsep{3mm}
  \scalebox{0.83}{
  \begin{tabular}{c|cccccc}
  \cline{1-6}
  Degree          &  Fan-in             & Pipeline   & $F\_max$      & \multicolumn{2}{c}{ Latency Results }  \\
          $D$              &  $F \times A$               & Strategy   & (MHz)$\uparrow$ & Clock Cycles$\downarrow$ & Latency(ns)$\downarrow$    \\ \hline
  \multirow{4}{*}{1}   & \multirow{2}{*}{4$\times$2} & (1)        & \bf{646}      & 6            & 9    \\
                           &                             & (2)        & 488           & \bf{3}       & \bf{6}    \\ \cline{2-6}
                           & \multirow{2}{*}{4$\times$3} & (1)        & \bf{571}      & 6            & 11    \\ 
                           &                             & (2)        & 363           & \bf{3}       & \bf{8}     \\ \hline\hline
  \multirow{4}{*}{2}   & \multirow{2}{*}{4$\times$2} & (1)        & \bf{568}      & 6            & 11    \\
                           &                             & (2)        & 440           & \bf{3}       & \bf{7}    \\ \cline{2-6}
                           & \multirow{2}{*}{4$\times$3} & (1)        & \bf{568}      & 6            & 11    \\ 
                           &                             & (2)        & 373           & \bf{3}       & \bf{8}    \\ \hline
  \end{tabular}}
  \egroup
\end{table}

We conducted additional experiments with PolyLUT-Add using the setup in Table~\ref{tb:setup_small}. This utilizes lower $F$ compared with the PolyLUT setup in Table~\ref{tb:setup}. $A=2$ is used for all models (which are denoted as ``HDR-Add2'', ``JSC-XL-Add2'', ``JSC-M Lite-Add2'', ``NID-Add2''). We also reduced the layer sizes in the DNN model for the UNSW-NB15 dataset. These configurations were found to reduce area whilst maintaining comparable accuracies. Optimization of these parameters may further improve results for specific applications.

Table~\ref{tb:comparison} shows the results and comparisons with prior works. Notably, PolyLUT applied $D=4$ for HDR, JSC-XL and NID Lite models and $D=6$ for the JSC-M Lite model, while PolyLUT-Add used smaller $D$. For comparable accuracy, the proposed PolyLUT-Add achieved a LUT reduction of 4.8$\times$, 6.5$\times$, 13.9$\times$ and 2.0$\times$ for the MNIST, JCS-XL, JSC-M Lite and UNSW-NB15 benchmarks respectively. 

Finally, we studied latency with comparable accuracy. Pipeline strategy-(2) in Figure~\ref{fg:clk} was used to minimize the number of clock cycles. Compared with PolyLUT, this approach achieved a 1.6$\times$, 1.3$\times$, 1.2$\times$, and 1.2$\times$ decrease for the four benchmarks, respectively. These significant reductions are attributed to lower polynomial degree $D$ and lower $F$.

\section{Conclusion}
\label{se:conclusion}

We introduced PolyLUT-Add, a novel technique designed to enhance connectivity between neurons in LUT-based networks to efficiently deploy DNNs at the edge. By combining base PolyLUT models, our approach mitigates scalability issues associated with conventional implementations and significantly improves efficiency. Specifically, we demonstrated that by utilizing a configuration of $A=2$, PolyLUT-Add with a lower polynomial degree $D$ and fan-in $F$ are sufficient to achieve comparable accuracy to PolyLUT. Over our benchmarks, PolyLUT-Add reduced LUT consumption by factors of 2.0-13.9 with a 1.2-1.6 times decrease in latency. The PolyLUT-Add architecture enhances LUT-based neural network performance in terms of area efficiency and latency. 

Future work could develop a targeted optimization technique for individually adjusting $F$, $A$, and $\beta$ parameters within each layer or neuron to substantially boost network accuracy, reduce latency, and optimize area efficiency.

\section{Acknowledgments}
Binglei Lou gratefully acknowledges financial support from Spatial Computational Learning.

\bibliographystyle{ieeetr}
\bibliography{ref}

\end{document}